\documentclass[conference]{IEEEtran}
\IEEEoverridecommandlockouts
\usepackage{cite}
\usepackage{amsmath,amssymb,amsfonts}
\usepackage{algorithmic}
\usepackage{graphicx}
\usepackage{textcomp}
\usepackage{subcaption}
\usepackage{graphicx}
\usepackage{xcolor}
\def\BibTeX{{\rm B\kern-.05em{\sc i\kern-.025em b}\kern-.08em
    T\kern-.1667em\lower.7ex\hbox{E}\kern-.125emX}}
\begin{document}

\title{Rusty Detection Using Image Processing For Maintenance Of Stations\\
{\footnotesize \textsuperscript{*}Detection of rust defects base on hsv color model}
}

\author{\IEEEauthorblockN{1\textsuperscript{st} Dao Duy Tung}
\IEEEauthorblockA{\textit{ViettelAI Research} \\
\textit{Viettel Network}\\
Ha Noi, Viet Nam \\
tungdd7@viettel.com.vn}
\and
\IEEEauthorblockN{2\textsuperscript{nd} Ho Xuan Hung}
\IEEEauthorblockA{\textit{ViettelAI Research} \\
\textit{Viettel Network}\\
Ha Noi, Viet Nam \\
hunghx@viettel.com.vn}
}

\maketitle

\begin{abstract}
This study addresses the challenge of accurately segmenting rusted areas on painted construction surfaces. A method leveraging digital image processing is explored to calculate the percentage of rust present on painted coatings. The proposed segmentation approach is based on the HSV color model. To equalize luminosity and mitigate the influence of illumination, a fundamental model of single-scale Retinex is applied specifically to the saturation component.

Subsequently, the image undergoes further processing, involving manual color filtering. This step is crucial for refining the identification of rusted regions. To enhance precision and filter out noise, the pixel areas selected through color filtering are subjected to the DBScan algorithm. This multi-step process aims to achieve a robust segmentation of rusted areas on painted construction surfaces, providing a valuable contribution to the field of corrosion detection and analysis.
\end{abstract}


\section{Introduction}
The external environment is the aggressive factor that
causes damage to the protective coatings of metal
constructions, tanks, power transmission lattice towers, etc.
This requires periodic monitoring of the coatings for the
detection of both corrosion damage and damage caused by
mechanical factors and human activities. It is necessary to
continuously monitor the safety of the use of such
constructions.

Automation of the process of diagnostics of surfaces of
metal structures involve the image acquisition and processing
of image series of the investigated object, detection and
localization of damage, determination of its type and degree,
formation of the protocol based on the results of the
diagnostics based on the relevant standards.

The problem of evaluating the degree of damage to the
surfaces of metal structures is difficult due to the presence of a
large number of objects of prolonged operation, direct access
to which is complicated or impossible. To assess such damage,
it is necessary to use appropriate equipment and highly skilled
specialists. At the same time, their work is associated with an
increased danger to health and life. In this regard, the
development of information technology and means of remote
visual control of surface damage of objects of long-term
operation is an urgent task of technical diagnostics.

Methods for automated evaluation of corrosion on metal
surfaces of digital images based on color, wavelet, and texture
analysis of damaged surfaces [1-4]. At the same time, for
solving such problems, neural networks were used, as well as
fuzzy logic, classification using vector support machines [4].

The authors [5] argue that clusters of two- and three-group
groups may incorrectly reflect the intensity of rust and rust
intensity on the rust image. They propose an approach to the
recognition of rust intensity for artificial neural networks. It
includes the standard deviation and the artificial neural
network to compose rust images, based on their intensity of
rust or stiffness. At the predetermined color spectrum
isolation, it can detect the intensity of rust in human
perception and play background noise.

To improve the accuracy of detection of rusted areas on
steel bridges Liao and Lee [6] proposed a recognition algorithm
of digital images, which consisted of three different detection
methods: the method of K-means for the layer H in color
space HSI, algorithm central dual-range (DCDR) in color
Space RGB and DCDR in the HSI color space.

Authors [7] also used the color space of HSI. Then the
decision tree J48 solution algorithm was used to classify the
rust area. They used images of steel bridges painted in
different colors. The authors achieved 97.48
identifying rusted areas for 119 test images.

Estimation of fractal parameters that correlate with two-dimensional and three-dimensional features of corrosive images was proposed in [8]. Valeti and Pakzad developed
detection of corrosion damage in power transmission lattice
towers using an unsupervised K-means clustering algorithm in a
Lab color space [9]. They segment input images into four
clusters and select clusters with corroded regions using hue
component from the HSV color model.

Development of new methods as part of intelligent systems
for the estimation of damage to protective coatings that reduce
risks to human life, minimize economic losses in monitoring
and service is an important task.

This article is organized as follows. In Part II, we provide a concise overview of the HSV color space, laying the foundation for subsequent discussions. Part III delves into the classic single-scale Retinex method, detailing its application within our proposed approach. Section IV elaborates on the color filtering method, accompanied by a small supporting tool. Moving on to Section V, we introduce the utilization of the DBScan algorithm, aimed at enhancing the clarity of rusty area identification within the images.

Finally, in Section VI, we present the outcomes of our experimentation. This structured arrangement aims to guide readers seamlessly through the exploration of HSV color spaces, the application of the single-scale Retinex method, the intricacies of color filtering, the efficacy of the DBScan algorithm, and, ultimately, the empirical results of our proposed methodology.

\section{HSV COLOR MODEL}
While color images are commonly represented in the RGB color model, alternative color spaces often prove more effective for image analysis. The HSV (Hue, Saturation, Value) color model, rooted in the human visual system, offers a more intuitive representation. It employs cylindrical coordinates to express RGB points, where H represents hue as an angular value, S denotes saturation, and V represents value.

In the HSV model, the hue values traverse the color spectrum, commencing with red at 0°, transitioning to green at 120°, blue at 240°, and cyclically returning to red at 360°. This unique representation aligns with human perception and facilitates a more nuanced analysis of color information within images.

Let’s assume that $H \in [0, 360]$, $S, V, R, G, B \in [0, 1]$, $Max = max\big\{R, G, B\big\}$, $Min = min\big\{R, G, B\big\}$, then:

\begin{equation}
H =\begin{cases}
0 & if ~Max = Min 
\\60 \frac{G-B}{Max-Min} & if ~Max = R~and~G \geq B
\\60 \frac{G-B}{Max-Min}+360 & if ~Max = R~and~G < B
\\60 \frac{B-R}{Max-Min}+120 & if ~Max = G
\\60 \frac{R-G}{Max-Min}+240 & if ~Max = B
\end{cases}
\end{equation}

\begin{equation}
S =\begin{cases} 
0 & if ~Max = 0 
\\1- \frac{Min}{Max} & otherwise
\end{cases}
\end{equation}

\begin{equation}
V = Max
\end{equation}

In this work, a set of test images of various objects was
used (see Fig. 1,2). The images were converted from RGB into
an HSV color model based on three color properties: Hue,
Saturation, and Value. Usually hue component is used in color
segmentation but unknown paint color and illumination level
can influence the results of segmentation. The saturation
component was the most relevant so it was used for further
processing (see Fig. 1)

\section{PROCESSING OF SATURATION COMPONENT}
The next step is the application of classic single-scale
retinex (SSR) to the saturation component of the input image. The
retinex is a model of human color vision presented by Land
[10] in 1964. The technology is based on the estimation and
normalization of illumination and works on the principle of
local contrast enhancement. It is suitable for processing
images containing dark or light low-contrast areas with
undistinguished local details.

Classic single-scale retinex algorithm [10] is basically the
reflectance function computation for each image pixel:

\begin{equation}
R(i,j)=\ln[L(i,j)]-\ln[L(i,j)*F(i,j)]=\ln\dfrac{L(i,j)}{L(i,j)*F(i,j)}
\end{equation}
where $L(i,j)$ is grayscale image intensity in pixel $(i,j)$, $i\in \overline{1,N}$, 
$j\in \overline{1,M}$, NxM is the image size, $F(i,j)$ is the Gaussian filter,
* is convolution of $F(i,j)$ with image L, that gives filtered image
\begin{equation}
\overline{L}(i,j)=L(i,j)*F(i,j)
\end{equation}
where $F(i,j)=Ke^{-(i^{2}+j^{2})/\sigma^{2}}$, $\sigma$ is a standard deviation,
K is a parameter calculated from equation:

\begin{equation}
K\sum_i\sum_jF(i,j)=1
\end{equation}

For the  single-scale retinex model (1) the final result of
processing of input image L is calculated by linear stretching
of $R(i,j) \in R$.

The parametric model of luminosity equalization using
SSR was presented in [11]. The parametric weighted contrast
was used instead of relative local contrast.

Next, to the enhanced saturation component, we use the
statistical method of Xu-segmentation [12] as the first iteration
of the ICM (Iterated conditional modes) method.

We calculate the threshold level $T^{*}$ by maximizing the
between-class variance:
\begin{equation}
	T^{*} = \underset{1\leq T<L}{argmax}\left\{\sigma_b^2(T)\right\}
\end{equation}
where $\sigma_b^2(T)=P_{0}(T)(\mu_{0}(T)-\mu)^{2}+P_{1}(T)(\mu_{1}(T)-\mu)^{2}$, $P_{0}(T)$, $P_{1}(T)$ are cumulative probabilities, $\mu_{0}(T)$, $\mu_{1}(T)$ are the mean values, $\sigma_b^2(T)$, $\sigma_1^2(T)$ are the variances of two classes.

Then the procedure was repeated within the segmented
class with smaller variance.

\section{COLOR FILTER WITH OPENCV}

The identification of rusty areas is a nuanced task, often distinguished by specific color characteristics, including shades of light red, orange, and yellow. To capture these subtleties, we employ OpenCV [14] in conjunction with a filtering mechanism based on the values of the three color channels: H (Hue), S (Saturation), and V (Value).

Recognizing the complexity of isolating rusty regions based on a single color range, we have developed a supplementary tool to enhance precision in the filtering process. This tool facilitates the meticulous selection of color codes, acknowledging that a comprehensive range is essential for accurately pinpointing diverse rust-induced hues. Through this iterative approach, we aim to elevate the accuracy and reliability of our color filtering technique, ensuring a meticulous synthesis that effectively captures the diverse spectrum of colors indicative of rust on surfaces.

The practical implementation involves two sets of images: the original images (a, b) and their corresponding masked counterparts, highlighting the rust-prone regions. The images at the top (a, b) depict the original photographs, while the lower set (c, d) introduces interactive trackbars designed for real-time manipulation.

The trackbars (c, d) are incorporated as a dynamic interface, allowing users to finely adjust parameters such as Hue (H), Saturation (S), and Value (V) through manual manipulation. These trackbars provide an intuitive means for users to observe immediate changes in the mask layer in response to alterations in the selected color ranges.

As users interact with the trackbars, the color filtering tool facilitates an instant feedback loop, enabling the precise calibration of the mask layer to accurately encapsulate the diverse hues indicative of rust. This interactive approach enhances the user's ability to fine-tune the filtering process, ensuring an adaptable and highly responsive mechanism for isolating rusted areas with meticulous precision.

\begin{figure}[htbp]
\centering
\begin{subfigure}{0.2\textwidth}
    \includegraphics[width=\textwidth]{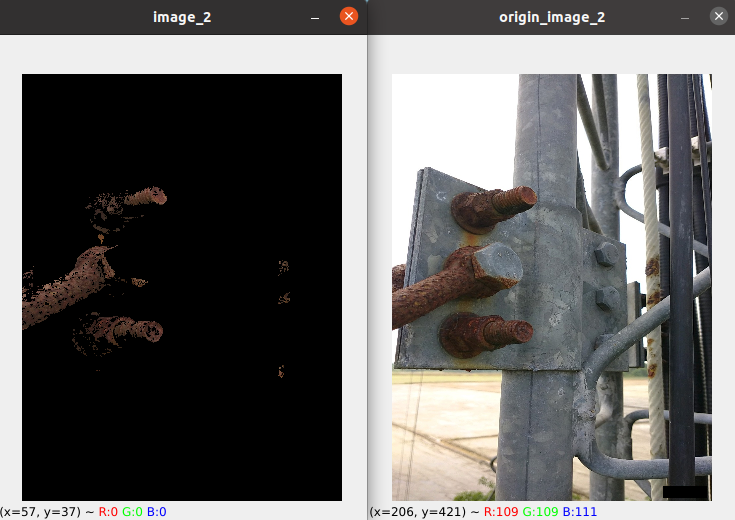}
    \caption{Image and mask 1}
    \label{fig: opencv_1}
\end{subfigure}
\hfill
\begin{subfigure}{0.2\textwidth}
    \includegraphics[width=\textwidth]{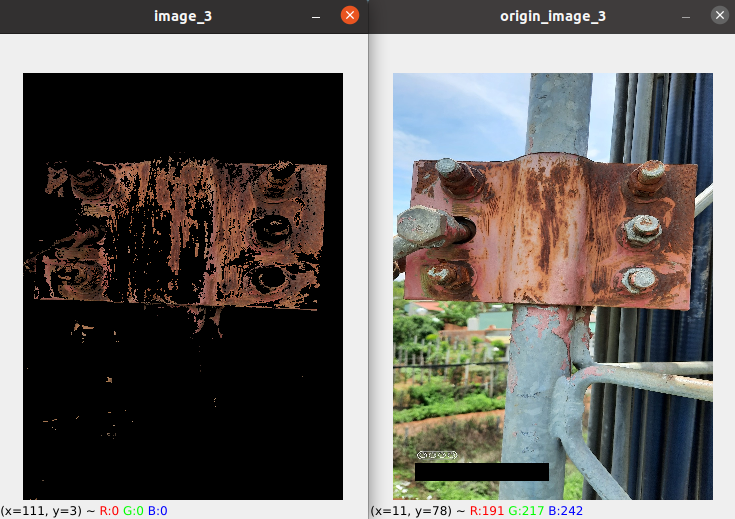}
    \caption{Image and mask 2}
    \label{fig: opencv_2}
\end{subfigure}
\hfill
\begin{subfigure}{0.2\textwidth}
    \includegraphics[width=\textwidth]{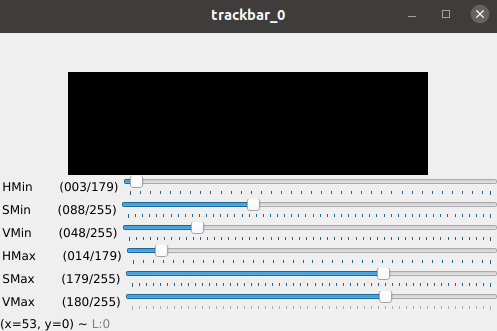}
    \caption{Trackbar 1}
    \label{fig: opencv_3}
\end{subfigure}
\hfill
\begin{subfigure}{0.2\textwidth}
    \includegraphics[width=\textwidth]{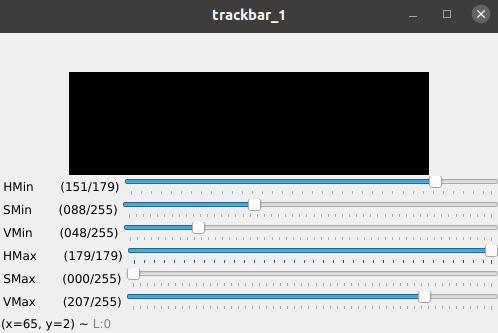}
    \caption{Trackbar 2}
    \label{fig: opencv_4}
\end{subfigure}
       
\caption{Color filter with OpenCV.}
\label{fig: opencv}
\end{figure}

\section{CLASSIC DBSCAN CLUSTERING METHOD}

Density-Based Spatial Clustering of Applications with Noise (DBSCAN). First introduced in [15], DBSCAN stands out as a prominent density-based clustering method renowned for its efficacy. Distinguishing itself from hierarchical and partition-based techniques, DBSCAN exhibits notable efficiency, particularly when confronted with clusters of arbitrary shapes, as illustrated in Figure 2.

At the core of DBSCAN lies the concept of reachability, gauging the number of neighbors within a specified radius for each data point [5]. This intrinsic characteristic facilitates the modeling of clusters with diverse and irregular shapes. In essence, DBSCAN allocates each point within the feature space to clusters that exhibit a substantial number of neighboring points in proximity. Conversely, points with low local densities, where their neighbors fall below the input radius, are designated as noise or outliers. This mechanism allows DBSCAN to robustly identify and delineate clusters in datasets characterized by irregular shapes and varying densities.

Although, DBSCAN does not require an a priori parameter k
as number of clusters, it requires two other parameters [5]:
\begin{itemize}
    \item $\epsilon$: radius, specifies how close two points should be one to another in order to be considered neighbors and belong to the same cluster.
    \item MinPts: minimum number of neighbors, specifies the minimum number of points within the $\epsilon$ radius in order to form a dense region.
\end{itemize}

\begin{figure}[htbp]
\centering{
    \includegraphics[width=.24\textwidth]{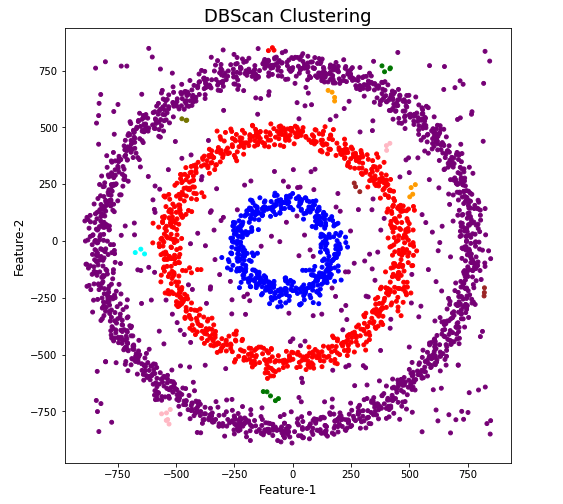}}
\caption{DBScan.}
\label{fig: dbscan}
\end{figure}

Following the application of color filtering, residual color clusters often persist, incorporating extraneous elements such as background features—examples being dry leaves, maintenance painting, and other non-relevant entities. In addition to isolating the rust-prone regions of an object, the filtering process may inadvertently include surrounding background regions.

To address this challenge and enhance the precision of rust identification, we integrate DBScan into our methodology. DBScan serves a dual purpose: firstly, as a noise reduction tool, and secondly, as a means to identify substantial rusted areas. The algorithm excels at clustering together regions of similar color, making it adept at discerning noise from significant clusters.

By employing DBScan, we effectively group clusters of color, distinguishing them from noise and irrelevant background elements. Notably, the algorithm aids in the identification of severe rusted areas, characterized by their expansive and contiguous nature. This two-fold application of DBScan significantly contributes to refining the accuracy of our rust detection process, ensuring a more precise delineation between relevant and extraneous color clusters.

\section{EXPERIMENT RESULT}

\begin{figure*}[ht]
\centering
\begin{subfigure}{0.135\textwidth}
    \includegraphics[width=\textwidth]{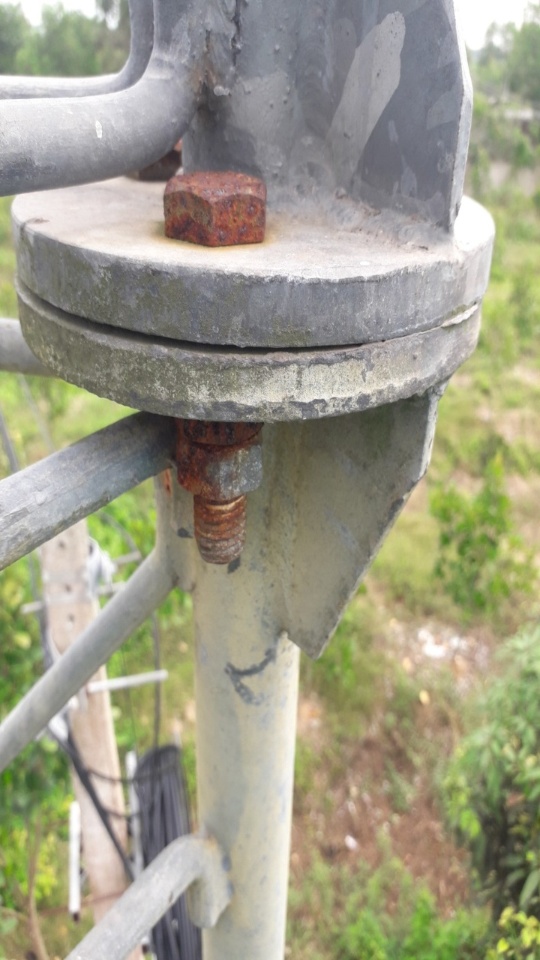}
    \caption{}
    \label{fig: exp_1}
\end{subfigure}
\hfill
\begin{subfigure}{0.135\textwidth}
    \includegraphics[width=\textwidth]{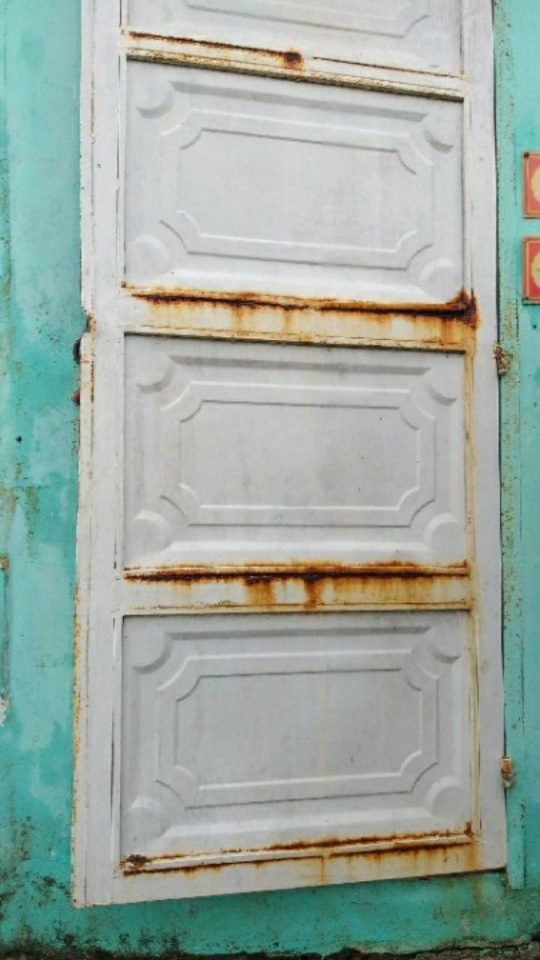}
    \caption{}
    \label{fig: exp_2}
\end{subfigure}
\hfill
\begin{subfigure}{0.135\textwidth}
    \includegraphics[width=\textwidth]{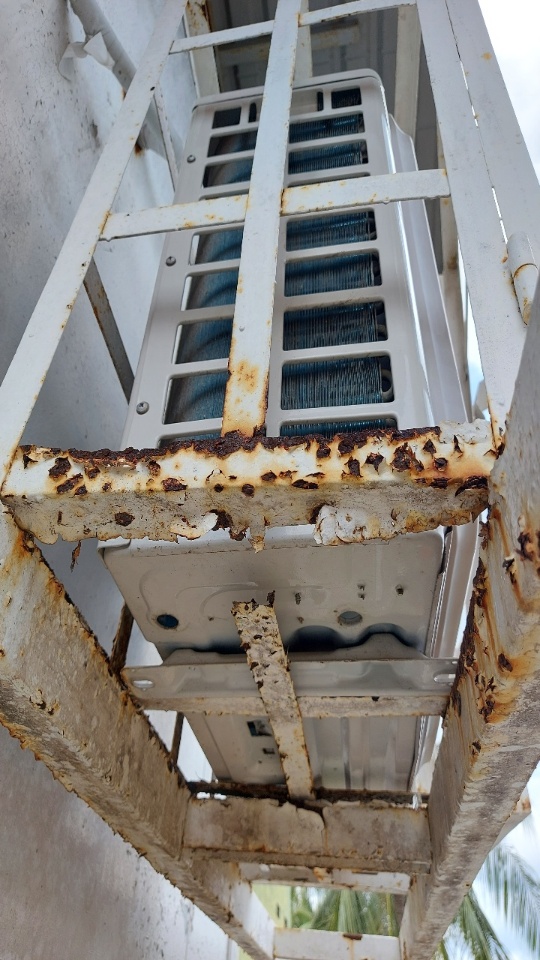}
    \caption{}
    \label{fig: exp_3}
\end{subfigure}
\hfill
\begin{subfigure}{0.135\textwidth}
    \includegraphics[width=\textwidth]{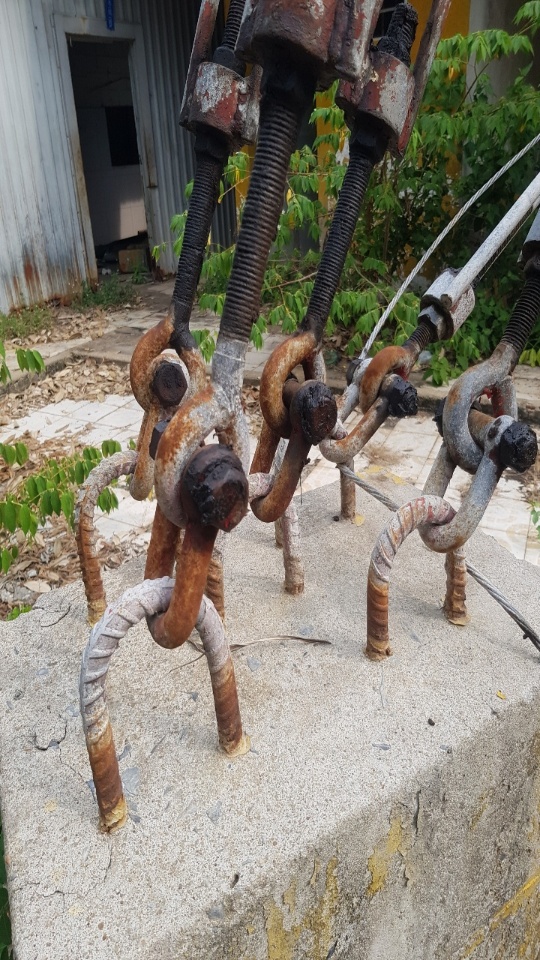}
    \caption{}
    \label{fig: exp_4}
\end{subfigure}
\hfill
\begin{subfigure}{0.135\textwidth}
    \includegraphics[width=\textwidth]{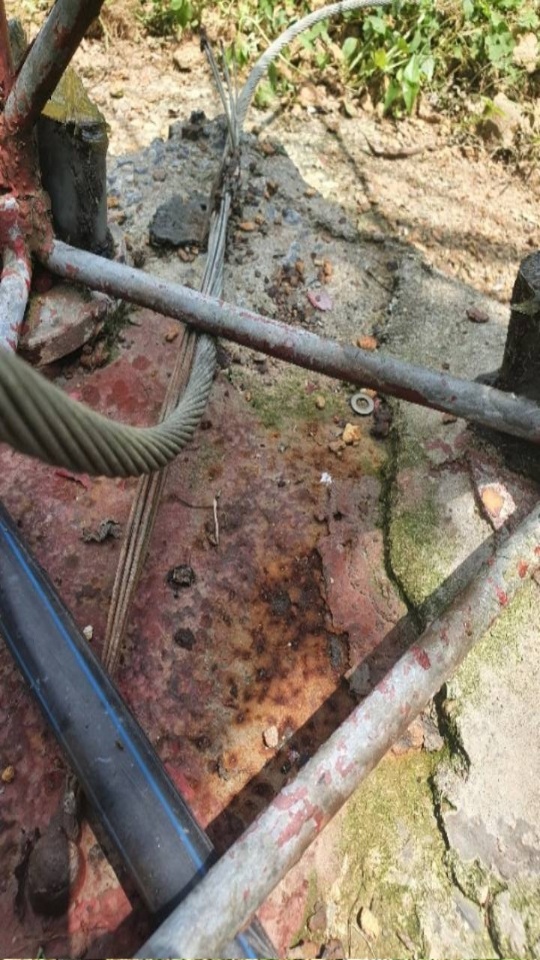}
    \caption{}
    \label{fig: exp_5}
\end{subfigure}
\hfill
\begin{subfigure}{0.135\textwidth}
    \includegraphics[width=\textwidth]{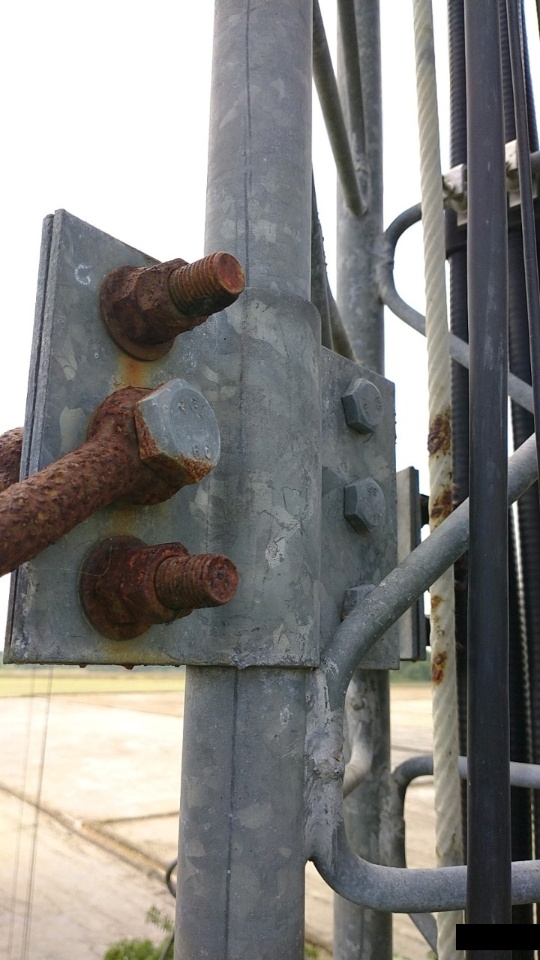}
    \caption{}
    \label{fig: exp_6}
\end{subfigure}
\hfill
\begin{subfigure}{0.135\textwidth}
    \includegraphics[width=\textwidth]{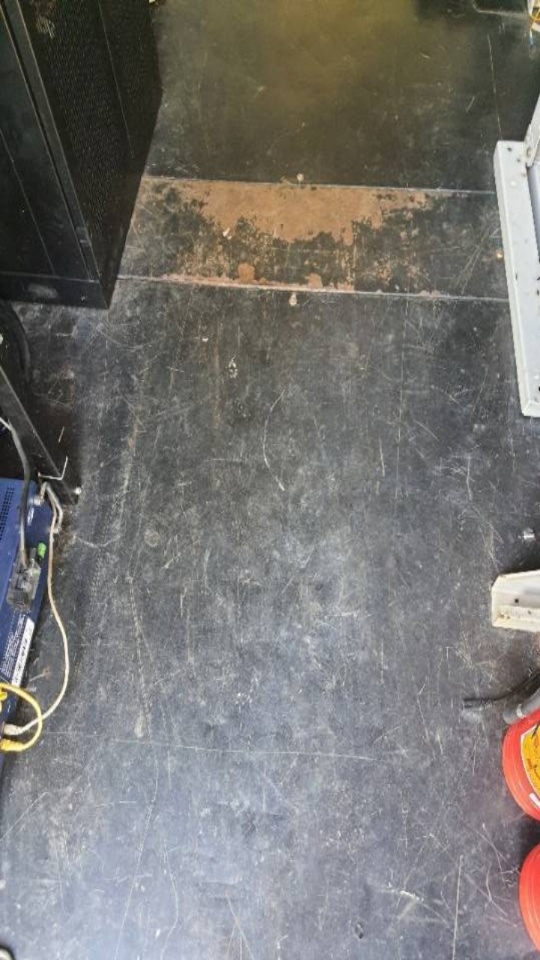}
    \caption{}
    \label{fig: exp_7}
\end{subfigure}
        
\caption{Objects of rusty detection.}
\label{fig:exp}
\end{figure*}

\begin{figure*}[ht]
\centering
\begin{subfigure}{0.135\textwidth}
    \includegraphics[width=\textwidth]{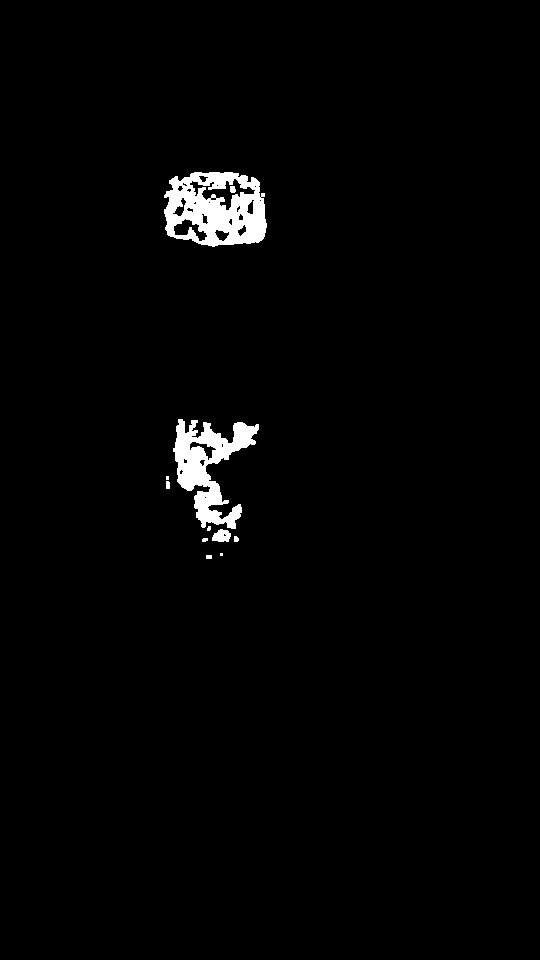}
    \caption{}
    \label{fig: mask_exp_1}
\end{subfigure}
\hfill
\begin{subfigure}{0.135\textwidth}
    \includegraphics[width=\textwidth]{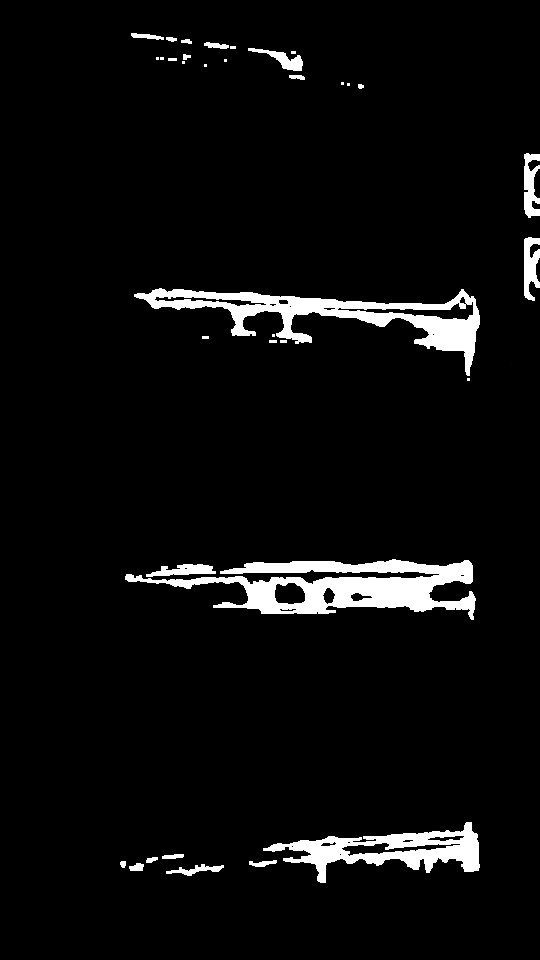}
    \caption{}
    \label{fig: mask_exp_2}
\end{subfigure}
\hfill
\begin{subfigure}{0.135\textwidth}
    \includegraphics[width=\textwidth]{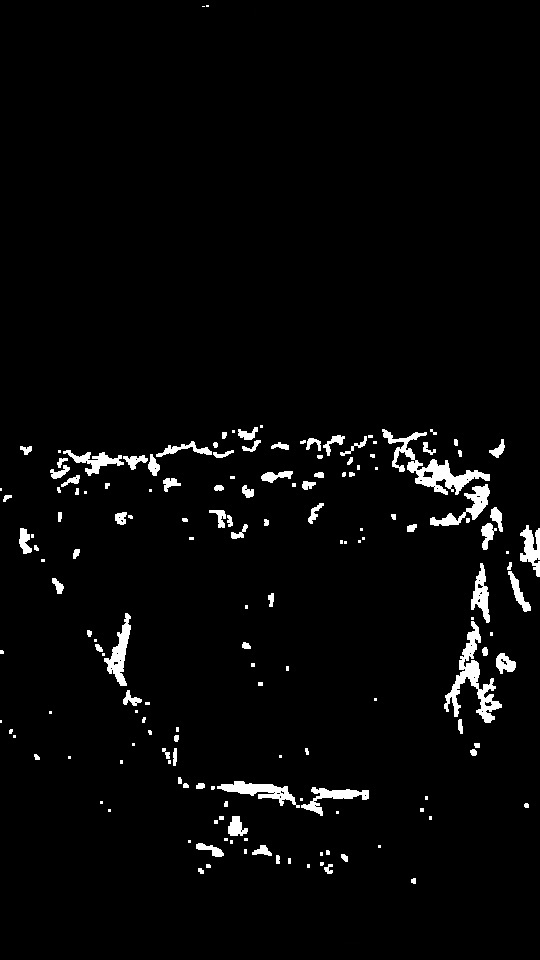}
    \caption{}
    \label{fig: mask_exp_3}
\end{subfigure}
\hfill
\begin{subfigure}{0.135\textwidth}
    \includegraphics[width=\textwidth]{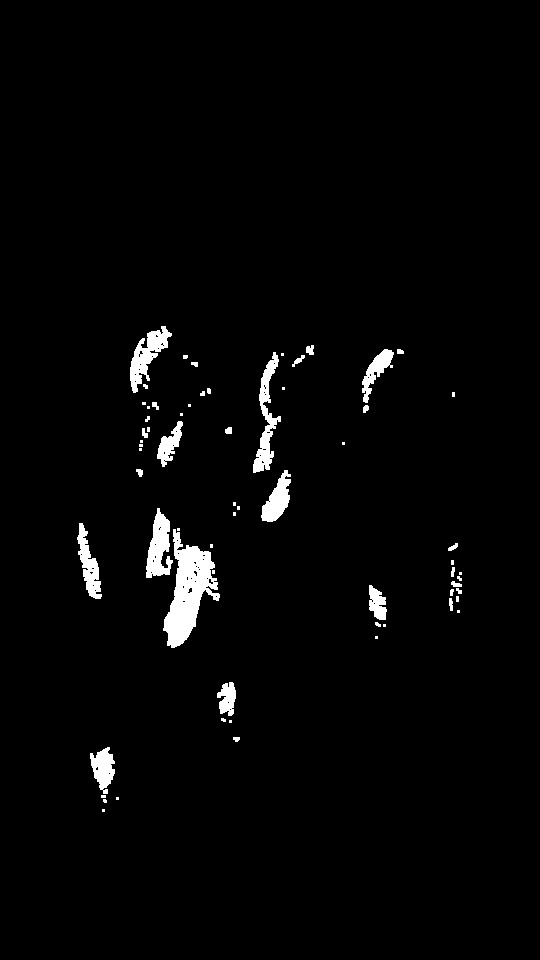}
    \caption{}
    \label{fig: mask_exp_4}
\end{subfigure}
\hfill
\begin{subfigure}{0.135\textwidth}
    \includegraphics[width=\textwidth]{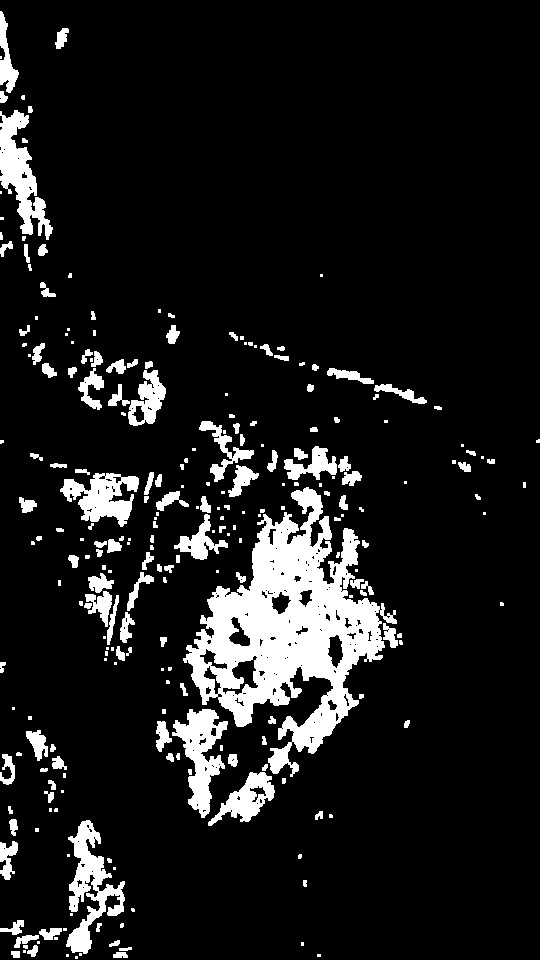}
    \caption{}
    \label{fig: mask_exp_5}
\end{subfigure}
\hfill
\begin{subfigure}{0.135\textwidth}
    \includegraphics[width=\textwidth]{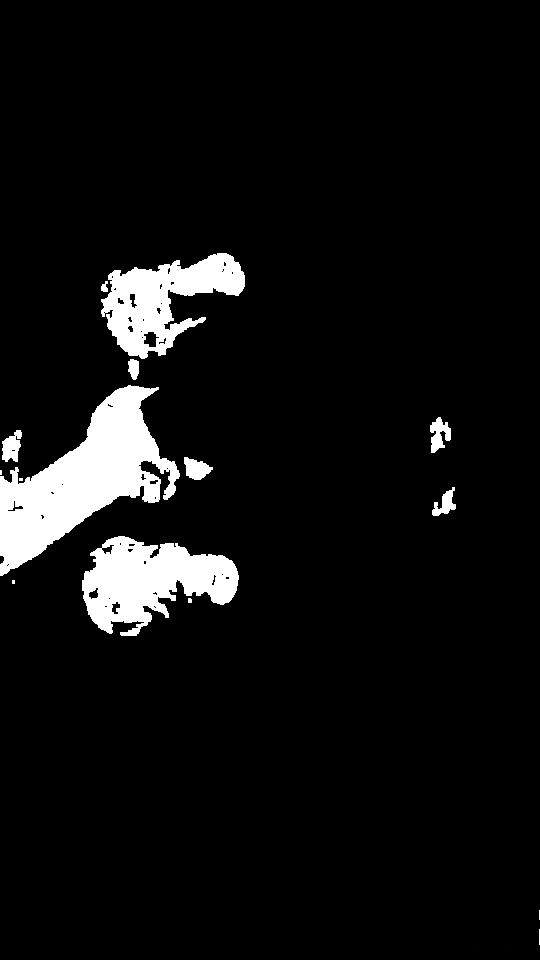}
    \caption{}
    \label{fig: mask_exp_6}
\end{subfigure}
\hfill
\begin{subfigure}{0.135\textwidth}
    \includegraphics[width=\textwidth]{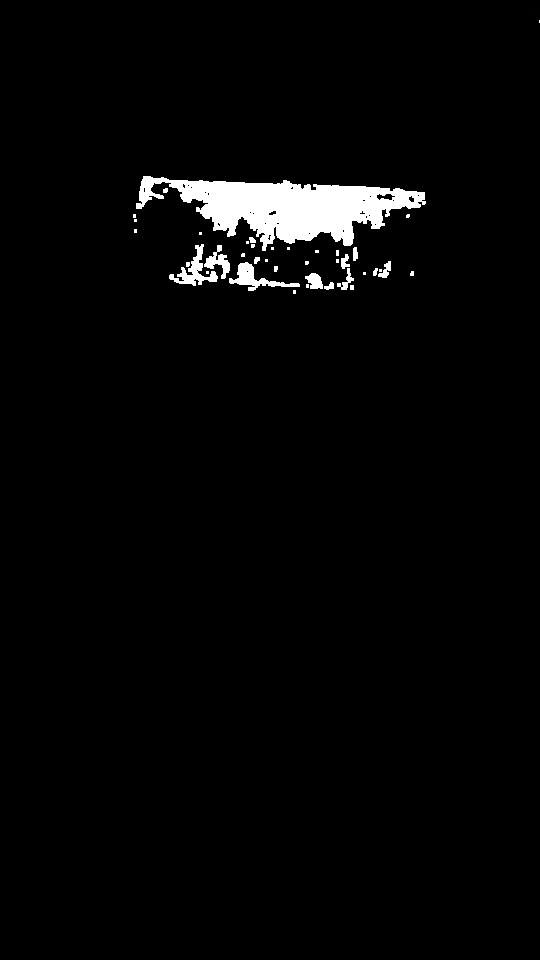}
    \caption{}
    \label{fig: mask_exp_7}
\end{subfigure}
        
\caption{Mask of rusty detection.}
\label{fig: mask_exp}
\end{figure*}

\begin{figure*}[ht]
\centering
\begin{subfigure}{0.135\textwidth}
    \includegraphics[width=\textwidth]{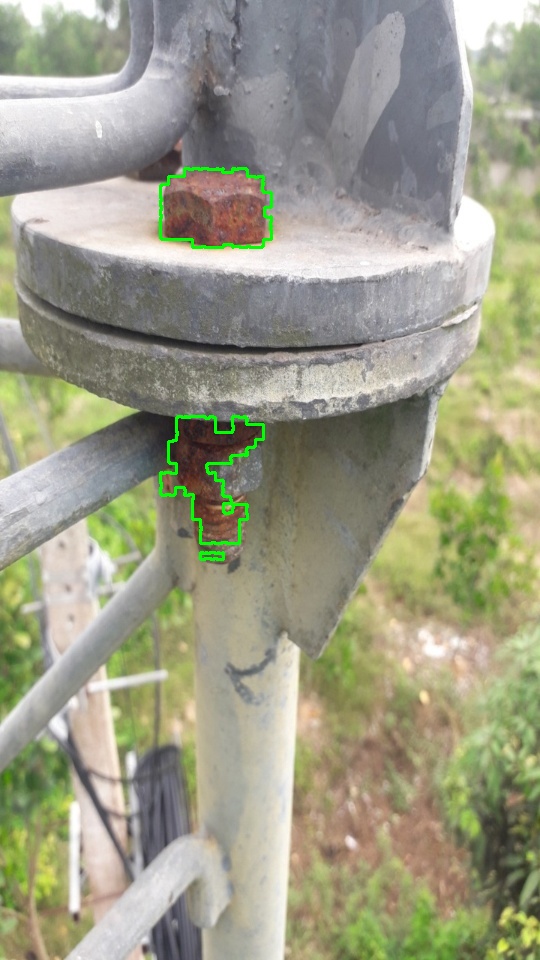}
    \caption{}
    \label{fig: exp_1_}
\end{subfigure}
\hfill
\begin{subfigure}{0.135\textwidth}
    \includegraphics[width=\textwidth]{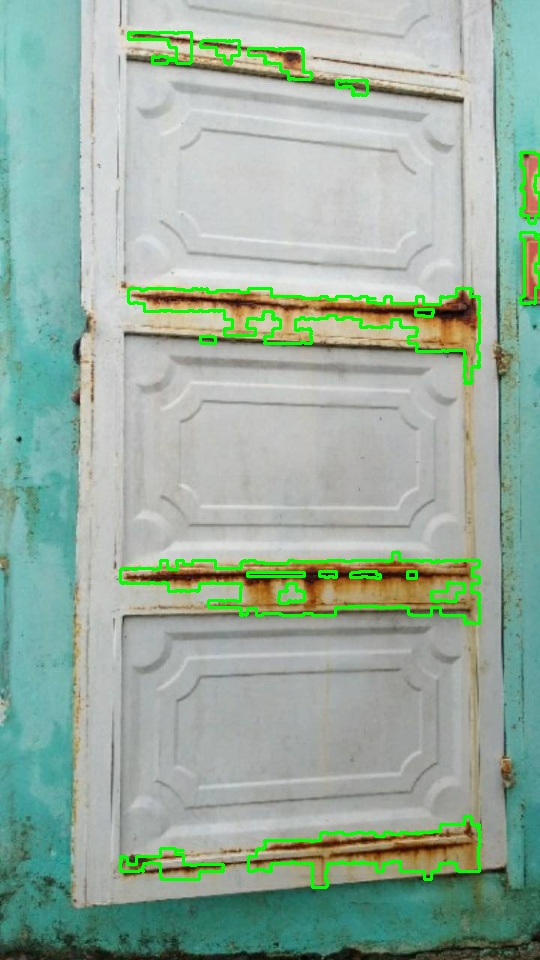}
    \caption{}
    \label{fig: exp_2_}
\end{subfigure}
\hfill
\begin{subfigure}{0.135\textwidth}
    \includegraphics[width=\textwidth]{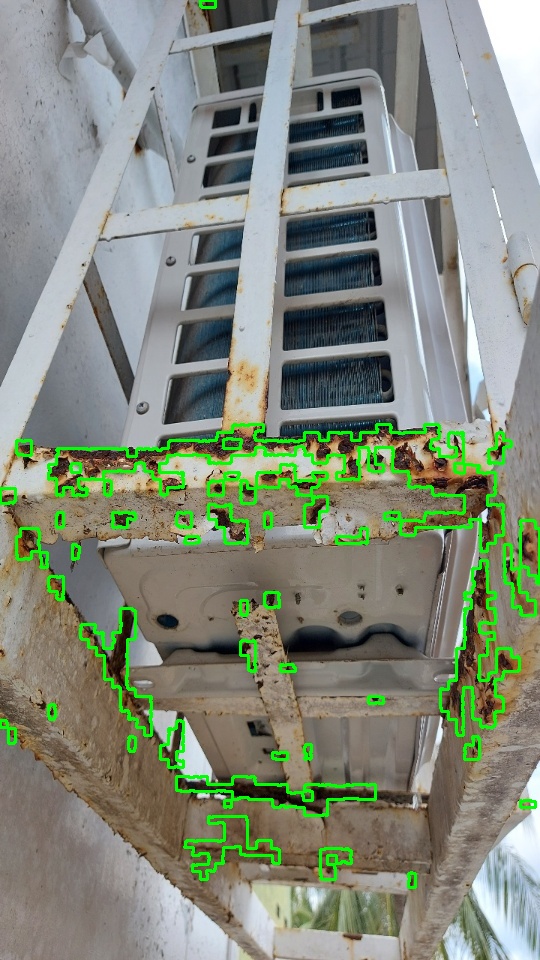}
    \caption{}
    \label{fig: exp_3_}
\end{subfigure}
\hfill
\begin{subfigure}{0.135\textwidth}
    \includegraphics[width=\textwidth]{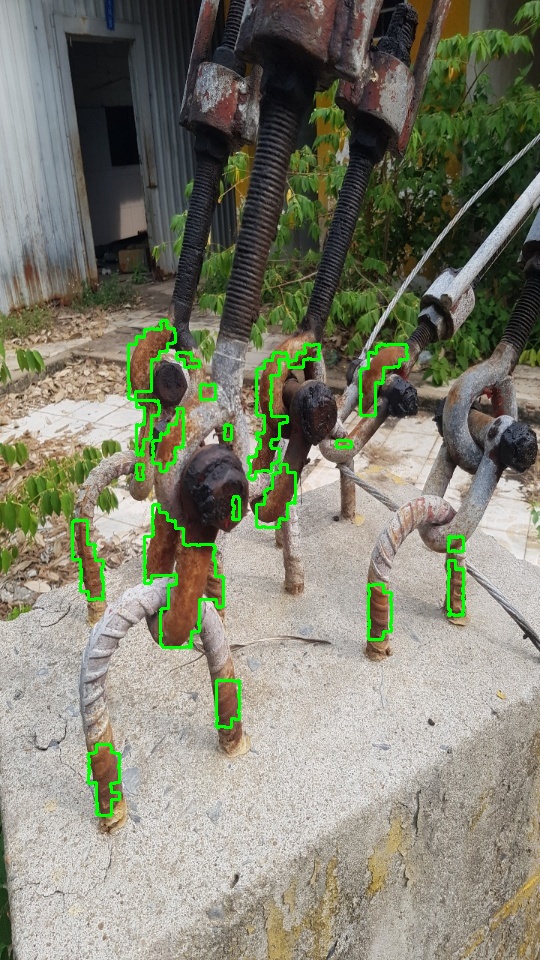}
    \caption{}
    \label{fig: exp_4_}
\end{subfigure}
\hfill
\begin{subfigure}{0.135\textwidth}
    \includegraphics[width=\textwidth]{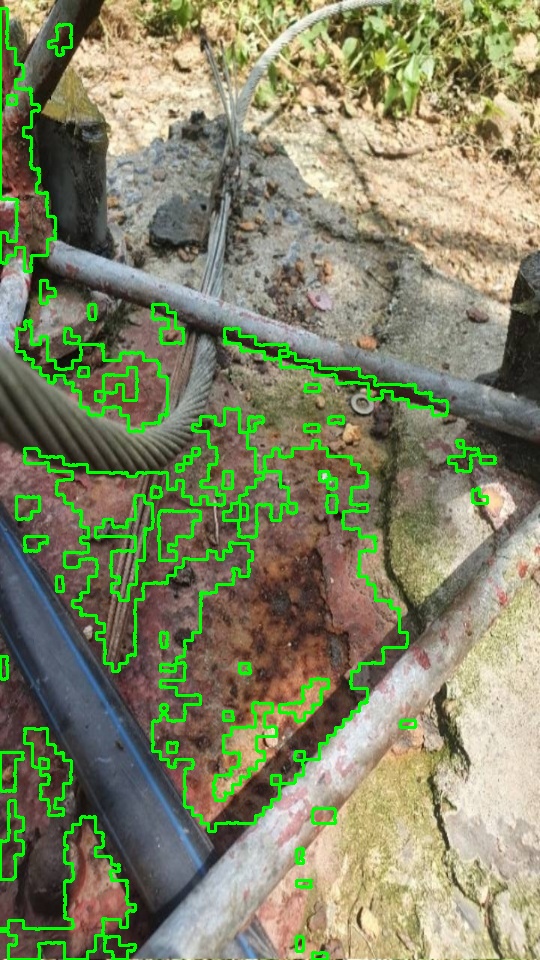}
    \caption{}
    \label{fig: exp_5_}
\end{subfigure}
\hfill
\begin{subfigure}{0.135\textwidth}
    \includegraphics[width=\textwidth]{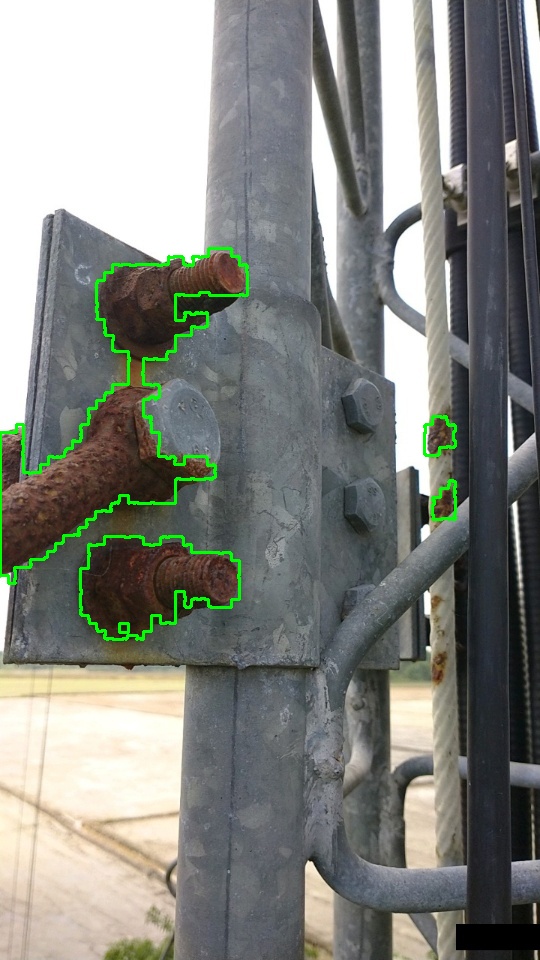}
    \caption{}
    \label{fig: exp_6_}
\end{subfigure}
\hfill
\begin{subfigure}{0.135\textwidth}
    \includegraphics[width=\textwidth]{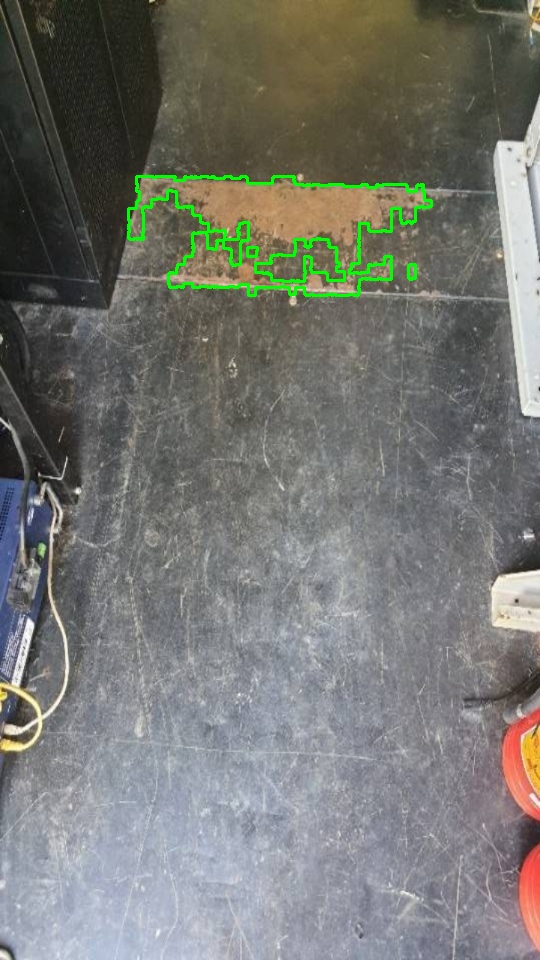}
    \caption{}
    \label{fig: exp_7_}
\end{subfigure}
        
\caption{Result of rusty detection..}
\label{fig: exp_figures_}
\end{figure*}

In the context of this research, the primary objective is to discern and evaluate rust through a classification model. The evaluation entails categorizing objects into two distinct classes: those exhibiting rust and those without. This process is conducted across a dataset comprising seven objects, each pertinent to station maintenance concerns.

The classification results are integral to gauging the model's efficacy in accurately identifying and distinguishing instances of rust. The seven objects chosen for experimentation encapsulate a spectrum of scenarios commonly encountered in station maintenance. The subsequent analysis and interpretation of the model's performance against these objects contribute crucial insights into the reliability and robustness of our proposed methodology. The visual representation of these results is presented below, providing a comprehensive overview of the classification outcomes.

\section{CONCLUSION}
In the context of this research, the primary objective is to discern and evaluate rust through a classification model. The evaluation entails categorizing objects into two distinct classes: those exhibiting rust and those without. This process is conducted across a dataset comprising seven objects, each pertinent to station maintenance concerns.

The classification results are integral to gauging the model's efficacy in accurately identifying and distinguishing instances of rust. The seven objects chosen for experimentation encapsulate a spectrum of scenarios commonly encountered in station maintenance. The subsequent analysis and interpretation of the model's performance against these objects contribute crucial insights into the reliability and robustness of our proposed methodology. The visual representation of these results is presented below, providing a comprehensive overview of the classification outcomes.


\begin{thebibliography}{00}
\bibitem{b1}R. Pidaparti, B. Hinderliter and D. Maskey, "Evaluation of Corrosion
Growth on SS304 Based on Textural and Color Features from Image
Analysis", ISRN Corrosion, vol. 2013, pp. 1-7, 2013.
\bibitem{b2}Shen, Heng-Kuang and Chen, Po-Han and Chang, Luh-Maan. (2013).
Automated Rust Defect Recognition Method Based on Color and
Texture Feature. Automation in Construction. 31. 338–356.
\bibitem{b3} M. Trujillo and M. Sadki, "Sensitivity analysis for texture models
applied to rust steel classification", Machine Vision Applications in
Industrial Inspection XII, 2004.
\bibitem{b4} F. Tsutsumi, H. Murata, T. Onoda, O. Oguri and H. Tanaka, "Automatic
corrosion estimation using galvanized steel images on power
transmission towers", 2009 Transmission and Distribution Conference and Exposition: Asia and Pacific, 2009.
\bibitem{b5} Heng-Kuang Shen, Po-Han Chen, Luh-Maan Chang, Human-visualperception-like intensity recognition for color rust images based on artificial neural network, Automation in Construction, Vol. 90, 2018, PP.
178-187, 10.1016/j.autcon.2018.02.023.
\bibitem{b6} Liao, Kuo-Wei and Lee, Yi-Ting. (2016). Detection of rust defects on
steel bridge coatings via digital image recognition. Automation in
Construction. 71. 10.1016/j.autcon.2016.08.008.
\bibitem{b7} Son, Hyojoo and Hwang, Nahyae and Changmin, Kim and Kim, Changwan.
(2014). Rapid and automated determination of rusted surface areas of a
steel bridge for robotic maintenance systems. Automation in
Construction. 42. 13–24. 10.1016/j.autcon.2014.02.016.

\bibitem{b8} S. Xu and Y. Weng, "A new approach to estimate fractal dimensions of
corrosion images", Pattern Recognition Letters, vol. 27, no. 16, pp.
1942-1947, 2006.
\bibitem{b9} Valeti, B.; Pakzad, S. Automated Detection of Corrosion Damage in
Power Transmission Lattice Towers Using Image Processing. In
Structures Congress 2017; American Society of Civil Engineers: Reston,
VA, USA, 2017; pp. 474–482.

\bibitem{b10} E. Land, "The Retinex", American Scientist., vol. 52, no. 2, pp. 247–
264, 1964.

\bibitem{b11} R. Vorobel, I. Ivasenko and O. Berehulyak, "Automatized computer
system for evaluation of rust using modified single-scale retinex, "2017
IEEE First Ukraine Conference on Electrical and Computer Engineering
(UKRCON), Kyiv, Ukraine, 2017, pp. 1002-1006.

\bibitem{b12} Xu, X., Xu, S., Jin, L., Song, E.: Characteristic analysis of Otsu
threshold and its applica-tions. Pattern Recognition Letters 32(7), 956-
961 (2011).

\bibitem{b13} Besag, J. E. (1986), "On the Statistical Analysis of Dirty Pictures",
Journal of the Royal Statistical Society, Series B, 48 (3): 259–302.

\bibitem{b14} R. Laganière. OpenCV 2 Computer Vision Application Programming Cookbook. Packt Publishing 2011.
\bibitem{b15} GRIDBSCAN: GRId Density-Based Spatial Clustering of Applications with Noise.

\end{thebibliography}
\end{document}